\documentclass[letterpaper, 10 pt, journal, twoside]{IEEEtran}

\usepackage{graphicx} % for pdf, bitmapped graphics files
\usepackage{epsfig} % for postscript graphics files
\usepackage{mathptmx} % assumes new font selection scheme installed
\usepackage{times} % assumes new font selection scheme installed
\usepackage{amsmath} % assumes amsmath package installed
\usepackage{amssymb}  % assumes amsmath package installed
\usepackage[mathscr]{eucal}
\usepackage{subfig}
\usepackage[noend]{algpseudocode}

\usepackage{algorithmicx,algorithm}
\usepackage{multirow} % Required for multirows
\usepackage{booktabs} % 导入三线表需要的宏包
\usepackage{array}
\usepackage{hyperref}
\usepackage[numbers,sort&compress]{natbib}
\usepackage{textcomp}

\usepackage{orcidlink}
\hypersetup{hypertex=true,
	colorlinks=true,
	linkcolor=blue,
	anchorcolor=blue,
	citecolor=blue}
\definecolor{lime}{HTML}{A6CE39}
\DeclareRobustCommand{\orcidicon}{
	\begin{tikzpicture}
		\draw[lime, fill=lime] (0,0)
		circle[radius=0.16]
		node[white]{{\fontfamily{qag}\selectfont \tiny \.{I}D}};
	\end{tikzpicture}
	\hspace{-2mm}
}
\foreach \x in {A, ..., Z}{%
	\expandafter\xdef\csname orcid\x\endcsname{\noexpand\href{https://orcid.org/\csname orcidauthor\x\endcsname}{\noexpand\orcidicon}}
}

\ifCLASSINFOpdf
\else

\fi

\usepackage{titlesec}
\titlespacing{\section}{0pt}{3.2ex plus .0ex minus .0ex}{1.05ex plus .0ex}
\titlespacing{\subsection}{0pt}{1.8ex plus .0ex minus .0ex}{0.7ex plus .0ex}
\UseRawInputEncoding
\begin{document}
	
	\title{Preparation of Papers for IEEE Robotics and Automation Letters (RA-L)}
	
	\markboth{IEEE Robotics and Automation Letters. Preprint Version. Accepted Month, Year}
	{FirstAuthorSurname \MakeLowercase{\textit{et al.}}: ShortTitle}

	\title{\huge DistillGrasp: Integrating Features Correlation with Knowledge Distillation for Depth Completion of Transparent Objects}
	
	\author{\IEEEauthorblockN{Yiheng Huang$^{\orcidlink{0009-0007-9527-6569}}$,
			Junhong Chen$^{\orcidlink{0000-0002-4874-9550}}$, 
			Nick Michiels$^{\orcidlink{0000-0002-7047-5867}}$, 
			Muhammad Asim$^{\orcidlink{0000-0002-6423-9809}}$\thanks{Y. Huang, J. Chen, M. Asim and W. Liu are with the College of Computer Science and Technology, Guangdong University of Technology, Guangzhou, 510006, China. E-mails: (huangyiheng.gdut@gmail.com, CSChenjunhong@hotmail.com, asimpk@gdut.edu.cn, liuwy@gdut.edu.cn.)  (Yiheng Huang and Junhong Chen contributed equally to this work.)    (Corresponding author: Junhong Chen and Wenyin Liu.)},  
			Luc Claesen$^{\orcidlink{0000-0003-0405-6290}}$, and Wenyin Liu$^{\orcidlink{0000-0002-6237-6607}}$\thanks{
				J. Chen and N. Michiels are with Hasselt University - Flanders Make, Expertise Centre for Digital Media, Hasselt, Belgium.}\thanks{
				L. Claesen is with Hasselt University, Hasselt, Belgium.
			}\thanks{ W. Liu is also with Zhongguancun Laboratory, Beijing, China.}}
		
	}
	
	% make the title area
	\maketitle 
	\pagestyle{empty}  % no page number for the second and the later pages
	\thispagestyle{empty} % no page number for the first page

	\begin{abstract}
		Due to the visual properties of reflection and refraction, RGB-D cameras cannot accurately capture the depth of transparent objects, leading to incomplete depth maps. To fill in the missing points, recent studies tend to explore new visual features and design complex networks to reconstruct the depth, however, these approaches tremendously increase computation, and the correlation of different visual features remains a problem. To this end, we propose an efficient depth completion network named DistillGrasp which distillates knowledge from the teacher branch to the student branch. Specifically, in the teacher branch, we design a position correlation block (PCB) that leverages RGB images as the query and key to search for the corresponding values, guiding the model to establish correct correspondence between two features and transfer it to the transparent areas. For the student branch, we propose a consistent feature correlation module (CFCM) that retains the reliable regions of RGB images and depth maps respectively according to the consistency and adopts a CNN to capture the pairwise relationship for depth completion. To avoid the student branch only learning regional features from the teacher branch, we devise a distillation loss that not only considers the distance loss but also the object structure and edge information. Extensive experiments conducted on the ClearGrasp dataset manifest that our teacher network outperforms state-of-the-art methods in terms of accuracy and generalization, and the student network achieves competitive results with a higher speed of 48 FPS. In addition, the significant improvement in a real-world robotic grasping system illustrates the effectiveness and robustness of our proposed system.
	\end{abstract}
	
	
	\IEEEpeerreviewmaketitle
	
	\begin{figure*}[htbp]
		
		\centering
		\includegraphics[width=0.93\linewidth]{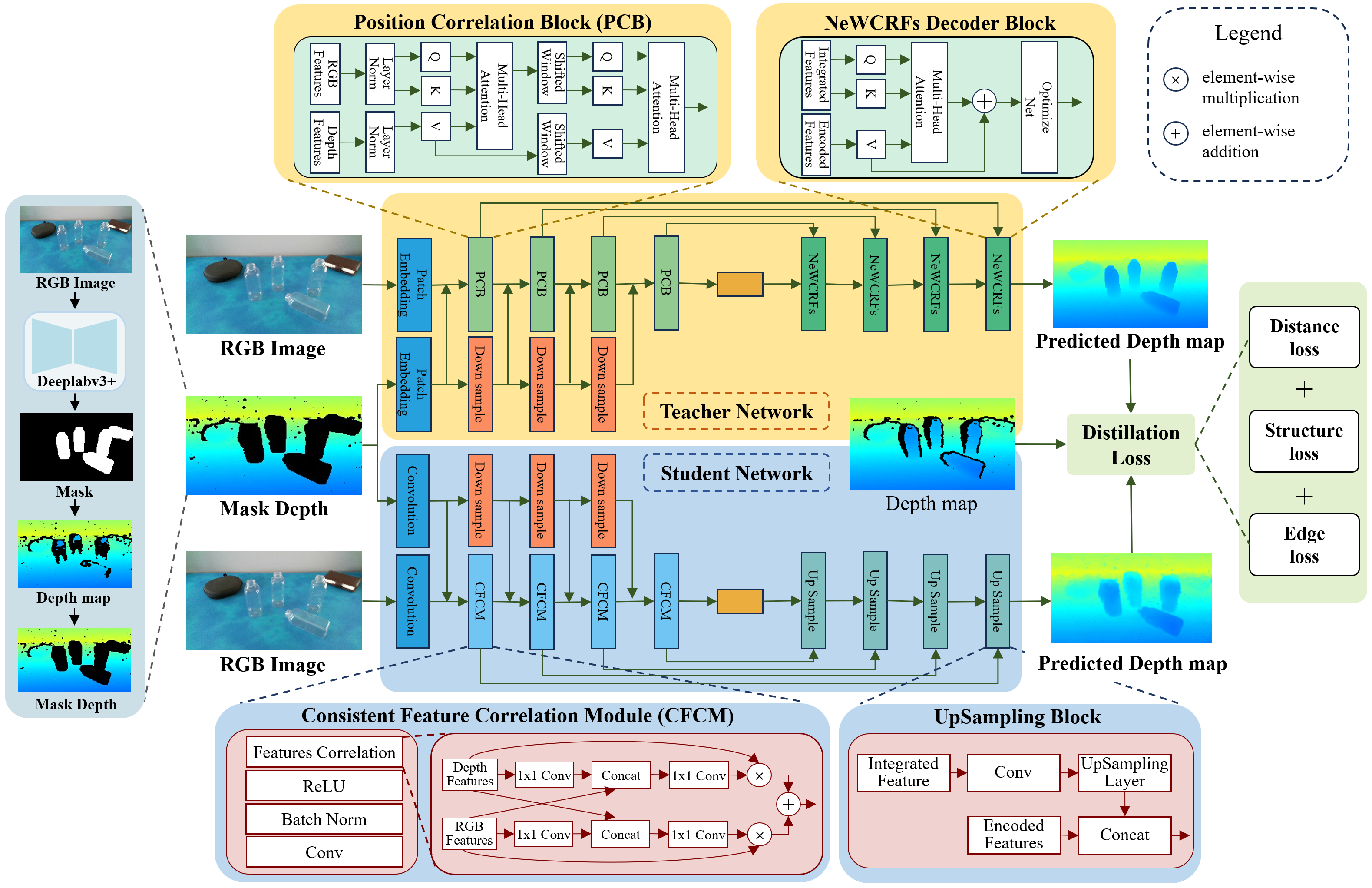}
		
		\captionsetup{font={small}}
		\caption{Overview of our proposed method DistillGrasp.}
		\label{Fig:1}
		\vspace{-3.4mm}
	\end{figure*}
	
	\section{Introduction}
	Transparent objects are widely used in various scenes, such as manufacturing, household services, etc. However, the perception of transparent objects has been recognized as a critical challenge \cite{ref1} due to their reflective and refractive surfaces that cannot be perceived by the depth cameras, resulting in inaccurate and missing depth values. Without accurate depth maps, many downstream robotics applications cannot be applied, e.g., object grasping, autonomous driving, human-robot interaction, etc.
	
	The task of the depth completion of transparent objects is to recover the complete depth maps from the input of RGB images and raw depth maps, which contains two aspects: correcting the drifting depth values caused by refraction and generating the missing depth values caused by reflection. To prevent the drifting depth values from affecting the learning of the network, authors in \cite{ref4, ref5} adopted a Deeplabv3+ network to remove the depth of transparent objects, and then extracted multiple visual maps as complementary information to complete the depth maps. However, the extraction of various visual maps not only increases the computational cost but also brings the problem of feature correlation. Instead of leveraging multiple visual maps, Chen et al. \cite{ref6} introduced Transformer to the depth completion of transparent objects in order to acquire contextual information for completing depth maps. Hong et al. \cite{ref7} integrated the DenseNet and Swin Transformer to extract both local and global features from RGB images and concatenated them with the depth maps to a U-Net architecture network for depth completion. Although these methods utilize advanced networks for fine-grained feature extraction, they still suffer from huge computational costs due to the complexity of the networks.
	
	In order to exploit the advantages of advanced networks while also maintaining model efficiency, in this paper, we propose a novel approach called DistillGrasp which preserves the advanced networks as the teacher networks to transfer the knowledge to the low-capacity student networks. Specifically, for the teacher network, we devise a position correlation block (PCB) which uses a transformer-based method to capture the structure information and establish the correspondence between RGB images and depth maps. For the student network, an efficient consistent feature correlation module (CFCM) is designed to capture the pairwise relationship based on reliable regions according to their consistency. To ensure that the student network can learn comprehensive features, a distillation loss incorporating distance, object structure, and edge information is introduced. Extensive experiments conducted on the ClearGrasp dataset verify the effectiveness and robustness of our methods. 
	
	In summary, our main contributions are as follows:
	\begin{itemize}
		\item  To the best of our knowledge, this is the first work on depth completion of transparent objects using knowledge distillation
		
		\item  For the teacher and student network, we separately devise two different correlation strategies named position correlation block (PCB) and consistent feature correlation module (CFCM) to capture the pairwise relationship, which guarantees accuracy and speed.
		
		\item  Considering the gap between the teacher and student network, we design a composite distillation loss consisting of distance loss, structural loss, and edge loss to ensure the student network can learn both local and global features.
		
		\item  Extensive experiments conducted on the ClearGrasp dataset illustrate that our teacher network outperforms state-of-the-art methods in accuracy and generalization, while the student network achieves competitive results with higher speed. The successful deployment of our system on a UR10e robot for grasping transparent objects verifies the effectiveness and robustness of our method.
	\end{itemize}

	\section{Related work}
	\subsection{Depth Completion for Transparent Object Perception}
	Depth completion for transparent objects aims to generate the missing depth values by referring to the existing RGB images and depth maps. Existing methods can be divided into two categories based on the number of viewpoints \cite{ref8}, the multi-view approaches and single-view approaches. The multi-view approaches complete the depth maps of transparent objects by utilizing information collected from different perspectives. For example, Klank et al. \cite{ref9} proposed to match two different views to locate the contradictory places, and then leveraged triangulation to correct the wrong point clouds. Authors in \cite{ref11, ref12} introduced the Neural Radiance Field (NeRF) to depth completion for transparent objects, but they suffered from huge computation costs. Compared with multi-view approaches, single-view approaches not only avoid capturing multiple views, but also reduce the time of processing large amounts of data from different perspectives. The primary single-view approaches are designed based on particular patterns \cite{ref13}, which cannot handle diverse scenarios and objects. To extend the application scenes, Sajjan et al. \cite{ref4} proposed an end-to-end network to extract multiple visual maps from a large amount of data to complete the missing depth. Hong et al. \cite{ref7} developed a U-Net network to complete the depth based on the correct depth filtered from the reflective and refractive regions. However, their methods focused on local features and ignored contextual information. Authors in \cite{ref6} extended the Transformer to depth completion for transparent objects, offering an enhanced perception of global information. Although the utilization of Transformer can improve accuracy, it also significantly increases computational costs, hindering the deployment in real-world environments.
	
	\subsection{Knowledge Distillation in Robotic Application}
	Knowledge distillation \cite{ref19} was first proposed to transfer refined knowledge from sophisticated teacher models to low-capacity student models, which has been widely deployed in various robotic applications, such as 3D object detection \cite{ref32, ref33}, scene segmentation \cite{ref34, ref35}, etc. Chen et al. \cite{ref32} proposed a cross-modal knowledge distillation approach that transferred the spatial information from LiDAR to multi-camera BEV for 3D object detection. Hong et al. \cite{ref33} applied knowledge distillation in monocular 3D object detection and extended it as a semi-supervised framework. Cen et al. \cite{ref34} designed a bidirectional fusion network to learn the enhanced 3D representation and leveraged a knowledge distillation framework for improving performance. Although a number of applications made in robotics, there are still relatively few works on robotic grasping. In this work, we introduce distillation learning to transparent object grasping, aiming to utilize the knowledge from advanced teacher networks to distill an efficient student network with high accuracy.
	\section{Method}
	\subsection{Overview}
	\vspace{0mm}
	As shown in Fig. \ref{Fig:1}, our proposed model contains a teacher network and a student network. In the teacher network, we devise an encoder-decoder network where the encoder leverages position correlation blocks (PCB) to establish the correspondence between RGB images and depth maps, while the decoder adopts NeWCRFs \cite{ref24} to restore the depth. In the student network, we use the same encoder-decoder network where the encoder utilizes Consistent feature correlation modules (CFCM) to capture pairwise relationships based on reliable positional information while the decoder uses the upsampling layers to recover the depth. Finally, a distillation loss with distance loss, structures loss, and edge loss is proposed to transfer the knowledge from the teacher network to the student network.
	
	\subsection{Teacher Network}
	The goal of our work is to improve the performance of synthetic data while keeping robustness to the real-world data. Thus, for the teacher network, we choose the Transformer network to extract spatial features since it performs better in capturing the overall structure of the objects. Specifically, we first adopt DeeplabV3+ to recognize transparent objects on RGB images and remove the corresponding areas on depth maps, which can avoid capturing wrong pairwise relationships between RGB images and depth maps. After that, patch embedding is employed on both channels respectively to split the images into patches and extract features. Because these features come from different channels and cannot provide useful mapping relationships for the network, we proposed our position correlation block to set up the correspondence between them.
	
	\textbf{Position correlation block (PCB).} Given a set of RGB image feature $I$ and depth feature $D$, we first initialize the features by utilizing a layer normalization:
	{
		\setlength\abovedisplayskip{0.2cm}
		\setlength\belowdisplayskip{0.2cm}
		\begin{gather}
			x_i  = LN(I) , x_d = LN(D)
		\end{gather}
	}where $LN$ refers to layer normalization. In order to correlate the two different features, we introduce the attention layer. Since there are incomplete points and missing points in the depth map, we choose RGB images as query $Q$ to search for the corresponding key $K$ in RGB images and value $V$ in the depth map. The attention layer can be denoted as follows:
	{
		\setlength\abovedisplayskip{0.2cm}
		\setlength\belowdisplayskip{0.2cm}
		\begin{gather}
			Q  = x_i , K = x_i , V=x_d  \notag \\%\notag????
			F_{qkv} = \text{SoftMax}(Q\cdot K^T+ B)\cdot V  + I  \\ %\notag????
			F_{corr} = MLP(LN(F_{qkv})) + F_{qkv}  \notag
		\end{gather}
	}where $B$ is the relative position bias, $MLP$ denotes a two-layer perceptron, $F_{qkv}$ and $F_{corr}$ represent the attention score and the correlated features. Although $F_{corr}$ establishes the correspondence between RGB images and depth maps, it only considers the correlation of the regional patches and ignores the context information. Inspired by the Swin Transformer \cite{ref26}, we adopt shifted window based attention to set up the connection between patches. Particularly, we apply the shifted window partitioning to the correlated features:
	{
		\setlength\abovedisplayskip{0.2cm}
		\setlength\belowdisplayskip{0.2cm}
		\begin{gather}
			I_{shift} = Shift(F_{corr}),D_{shift} = Shift(x_d) 
		\end{gather}
	}where $Shift$ denotes shifted window partitioning, $I_{shift}$ and $D_{shift}$ represent the shifted RGB image features and the shifted depth features. Similar to the previous attention layer, the shifted image features are used to search for the corresponding key $K$ in shift RGB image features and value $V$ in the shifted depth features. Finally, a patch merging layer is adopted to reduce the height and width by half and expand the number of channels to twice. By stacking up four sets of PCBs, we can obtain the encoding features.
	
	In terms of the decoding stage, we follow \cite{ref24} which develops a bottom-up structure that consists of four neural windows fully-connected Conditional Random Fields modules (NeWCRFs), where each module receives the features from the previous layer as well as the low-level features from the encoder through skip connections and outputs the upscale features utilizing a shuffle operation. Finally, the decoder outputs the completed depth map.
	\subsection{Student Network}
	The student network aims to keep pace with the teacher network while staying efficient at the same time. Therefore, for the student network, we choose CNN as our feature extractor because of its efficiency. Specifically, we first remove the depth of transparent objects by leveraging DeeplabV3+. Different from the teacher network that leverages RGB images to query corresponding positions on depth maps, our student network utilizes a more direct and efficient module called consistent feature correlation module which establishes correspondence based on the reliable positions of the RGB images and depth maps. The detail of the module is introduced as follows.
	
	\textbf{Consistent feature correlation module (CFCM).} Given a set of RGB image feature $I$ and depth feature $D$, we set up two branches to calculate the consistency respectively, where each branch has the same structure except for the input. For clarity, we take the RGB image branch as an example. In particular, we first leverage a $1  \times 1$ convolution network to aggregate the different channel information and increase the number of channels. To avoid using too many convolutional layers that destroy the original spatial structure, the aggregated features are concatenated with the original depth features and sent into a $1  \times 1$ convolution network to compute the consistency score $C_I$. The consistency score $C_I$ is formulated as follows:
	{
		\setlength\abovedisplayskip{0.2cm}
		\setlength\belowdisplayskip{0.2cm}
		\begin{equation}
			\begin{split}
				C_I = Conv_{1 \times 1}( Conv_{1\times1}(I) \oplus D )
			\end{split}
		\end{equation}
	}where $ Conv_{1\times1}$ represents the  $1  \times 1$ convolution network,$ \oplus $ indicates the channel concatenation, By multiplying the original RGB image feature $I$ and the consistency score $C_I$, we can obtain the reliable RGB image and $P_I$.
	{
		\setlength\abovedisplayskip{0.2cm}
		\setlength\belowdisplayskip{0.2cm}
		\begin{equation}
			\begin{split}
				P_I = C_I\odot I 
			\end{split}
		\end{equation}
	}where $ \odot $ denotes element-wise multiplication. Following the same process, we can obtain a reliable depth map $P_D$. These two reliable visual maps are combined through element-wise addition to generate the integrated features $F$.
	{
		\setlength\abovedisplayskip{0.2cm}
		\setlength\belowdisplayskip{0.2cm}
		\begin{equation}
			\begin{split}
				F = P_I + P_D 
			\end{split}
		\end{equation}
	}Finally, we devise an efficient CNN block to capture the pairwise relationships, where each block consists of a ReLU activation, a batch normalization, and a convolution layer to capture pairwise relationships. After stacking up four sets of CFCM, we can obtain the decoded features.
	
	In terms of the decoder, four upsampling blocks are deployed, where each block integrates a CNN network and an upsampling layer. Similarly, the encoded features are transferred to the decoder via skip connections, ensuring the multiscale features can be fully utilized. Finally, the decoder outputs the completed depth map.

	\subsection{Distillation Loss}
	Considering that the CNN-based student network can not acquire sufficient context features and leads to an unsatisfactory performance on the overall completion, we design our distillation loss into three components: distance loss, structural loss, and edge loss, where the distance loss is used to shrink the distance error, and the structural loss and edge loss ensure the overall structure of the completed depth.
	
	\textbf{Distance Loss.} Scale-Invariant Logarithmic (SILog) loss \cite{ref27} has been proven to have the ability to capture scale-invariant depth information, thus we introduce it as our distance loss. Particularly, we first calculated the logarithm difference between the predicted depth map $\Delta E_p$ and the real depth $\Delta E_{gt}$:
	{
		\setlength\abovedisplayskip{0.2cm}
		\setlength\belowdisplayskip{0.2cm}
		\begin{equation}
			\begin{split}
				\Delta E_p =logD_{s} - logD_{t} , \Delta E_{gt} =logD_s - logD_{gt}
			\end{split}
		\end{equation}
	}where $D_s$ and $D_{t}$ denote the predicted depth map completed by the student network and teacher network respectively. $D_{gt}$ is the groundtruth depth value. For K valid values in the depth map, the SILog loss $\mathscr{L}_d$ is computed as follows:
	{
		\setlength\abovedisplayskip{0.2cm}
		\setlength\belowdisplayskip{0.2cm}
		\begin{equation}
			\begin{split}
				\mathscr{L}_d &= \alpha \sqrt{\frac{1}{K} \sum_{}^{}\Delta E_{gt}^2 -  \frac{\lambda}{K^2} (\sum_{}^{}\Delta E_{gt})^2} \\
				& + \beta \sqrt{\frac{1}{K} \sum_{}^{}\Delta E_{p}^2 -  \frac{\lambda}{K^2} (\sum_{}^{}\Delta E_{p})^2}
			\end{split}
		\end{equation}
	}where $\alpha$ and $\beta$ are scale constants and $\lambda$ is the variance minimizing factor. In our experiments, $\alpha$ and $\beta$ are set to 3 and 7 respectively, $\lambda$ is set to 0.85.
	
	\begin{figure*}[htbp]
		\centering
		\includegraphics[width=1\linewidth]{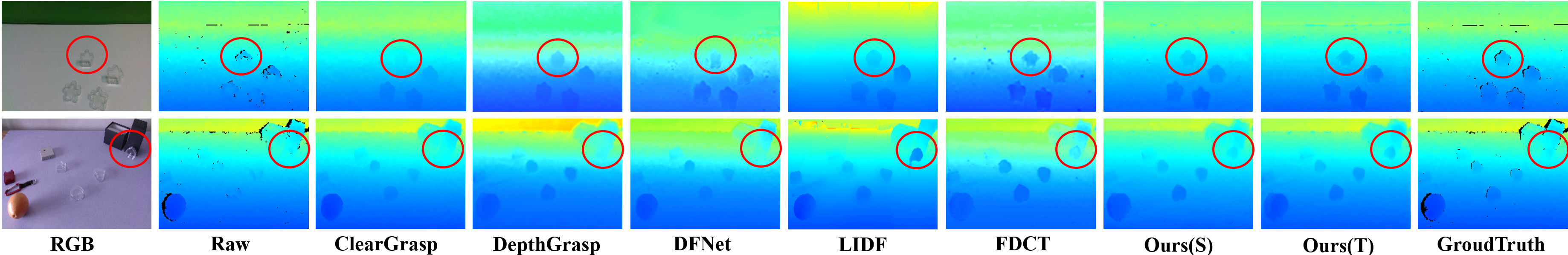}
		
		\captionsetup{font={small}}
		\caption{Qualitative comparison of the state-of-the-art approaches, where the generated details are highlighted with red circles.}
		\vspace{-4.0mm}
		\label{Fig:3}
	\end{figure*}
	
	\textbf{Structural Loss.} We utilize the structural metrics from the structural similarity index measure (SSIM) and calculate their mean square error (MSE) as our structural loss. Specifically, the structural metric is defined as follows:
	{
		\setlength\abovedisplayskip{0.2cm}
		\setlength\belowdisplayskip{0.2cm}
		\begin{equation}
			\begin{split}
				C(x,y) = \frac{\sigma_{x,y}+\theta}{\sigma_{x} \sigma_{y}+ \theta}
			\end{split}
		\end{equation}
	}where $x$ and $y$ represent the depth maps, $\sigma_{x}$ is the variance of the depth maps, $\sigma_{x,y}$ is the covariance of the depth maps, and $\theta$ is the constant to prevent the dominator from being zero. In our work, we compute two sets of structural metrics, where the first metric $C(D_{gt},D_s)$ is computed between the groundtruth depth maps and the predicted depth maps from the student network, and the other metric $C(D_t,D_s)$ is calculated between the predicted depth maps from the teacher network and the student network. Finally, the structural loss $\mathscr{L}_{s}$ is computed as follows:
	{
		\setlength\abovedisplayskip{0.2cm}
		\setlength\belowdisplayskip{0.2cm}
		\begin{equation}
			\begin{split}
				\mathscr{L}_{s} = \text{MSE}(C(D_{gt},D_t),C(D_t,D_s))
			\end{split}
		\end{equation}
	}where $\text{MSE}$ denotes the mean square error.
	
	\textbf{Edge Loss.} Since the boundary of the transparent objects is always confused with the background, our edge loss is designed to capture the horizontal and vertical variations of the consecutive points. Particularly, the horizontal variations are defined as follows:
	{
		\setlength\abovedisplayskip{0.2cm}
		\setlength\belowdisplayskip{0.2cm}
		\begin{equation}
			\begin{split}
				V_x^r = |D_{x,y}^r - D_{x+1,y}^r| \odot M_{x} 
			\end{split}
	\end{equation}}where $V_x^r$ denotes the horizontal variations of the depth maps $r$, $D_{x,y}^r$ represents the depth maps $r$, $M_{x}$ is the mask of the transparent objects in the horizontal direction. Similarly, the vertical variations are defined as follows:
	{
		\setlength\abovedisplayskip{0.2cm}
		\setlength\belowdisplayskip{0.2cm}
		\begin{equation}
			\begin{split}
				V_y^r = |D_{x,y}^r - D_{x,y+1}^r| \odot M_{y} 
			\end{split}
		\end{equation}
	}where $V_y^r$ denotes the vertical variations of the depth maps $r$ and $M_{y}$ is the mask of the transparent objects in the vertical direction. In our work, we compute two edge losses, where the first loss $\mathscr{L}_e^{gt,s}$ is computed between the groundtruth depth maps and the predicted depth maps from the student network.
	{
		\setlength\abovedisplayskip{0.2cm}
		\setlength\belowdisplayskip{0.2cm}
		\begin{equation}
			\begin{split}
				\mathscr{L}_e^{gt,s} = \text{MSE}(V_x^gt,V_x^s) + \text{MSE}(V_y^gt,V_y^s)
			\end{split}
		\end{equation}
	}and the other $\mathscr{L}_e^{t,s}$ is calculated between the predicted depth maps from the teacher network and the student network.
	{
		\setlength\abovedisplayskip{0.2cm}
		\setlength\belowdisplayskip{0.1cm}
		\begin{equation}
			\begin{split}
				\mathscr{L}_e^{t,s} = \text{MSE}(V_x^t,V_x^s) + \text{MSE}(V_y^t,V_y^s)
			\end{split}
	\end{equation}}
	
	In summary, the total loss of DistillGrasp $\mathscr{L}$ is defined as follows:
	{
		\setlength\abovedisplayskip{0.2cm}
		\setlength\belowdisplayskip{0.2cm}
		\begin{equation}
			\begin{split}
				\mathscr{L} = \mathscr{L}_{d} + \lambda_1 *  \mathscr{L}_{s} + \lambda_2 *  \mathscr{L}_e^{gt,s} + \lambda_3 *   \mathscr{L}_e^{t,s}  
			\end{split}	
		\end{equation}
	}where $\lambda_1, \lambda_2$ and $\lambda_3$ are empirically set as 0.1, 0.3, and 0.7, respectively.
	
	\section{EXPERIMENTS}
	\subsection{Dataset}
\begin{figure}[htbp]
		\vspace{1.0mm}
		\centering
		\subfloat[Syntheic images]{
			\includegraphics[width=0.98\linewidth]{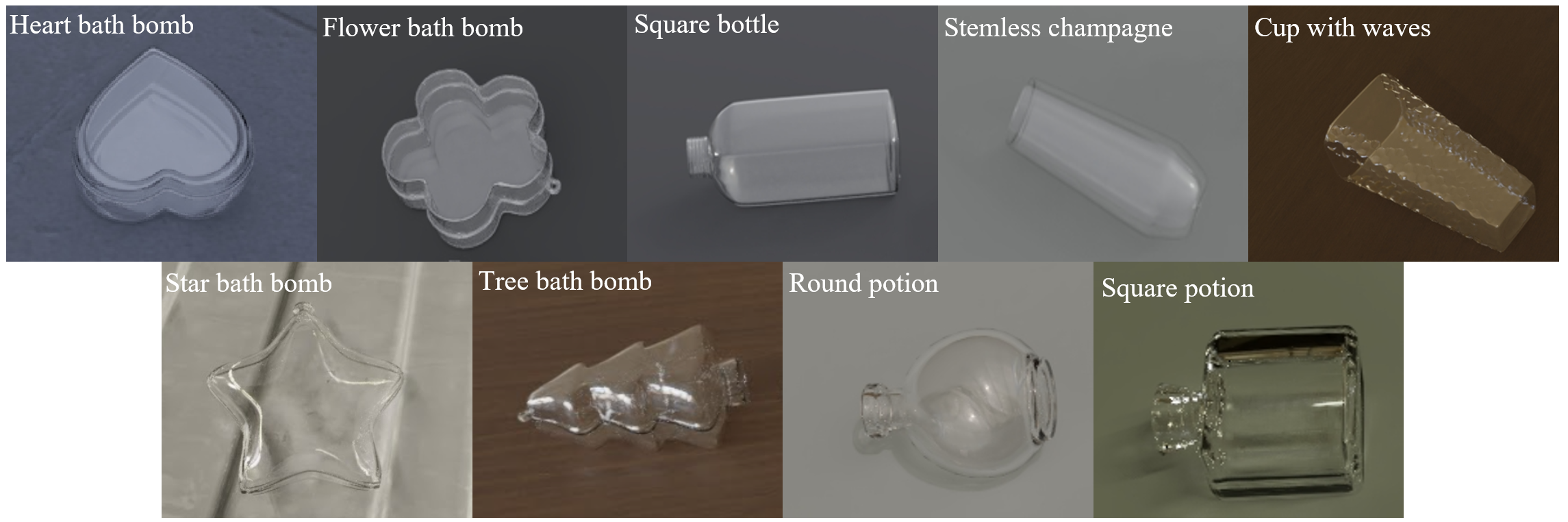}}
		\vspace{-2.5mm}
		\subfloat[Real-world images]{
			\includegraphics[width=0.98\linewidth]{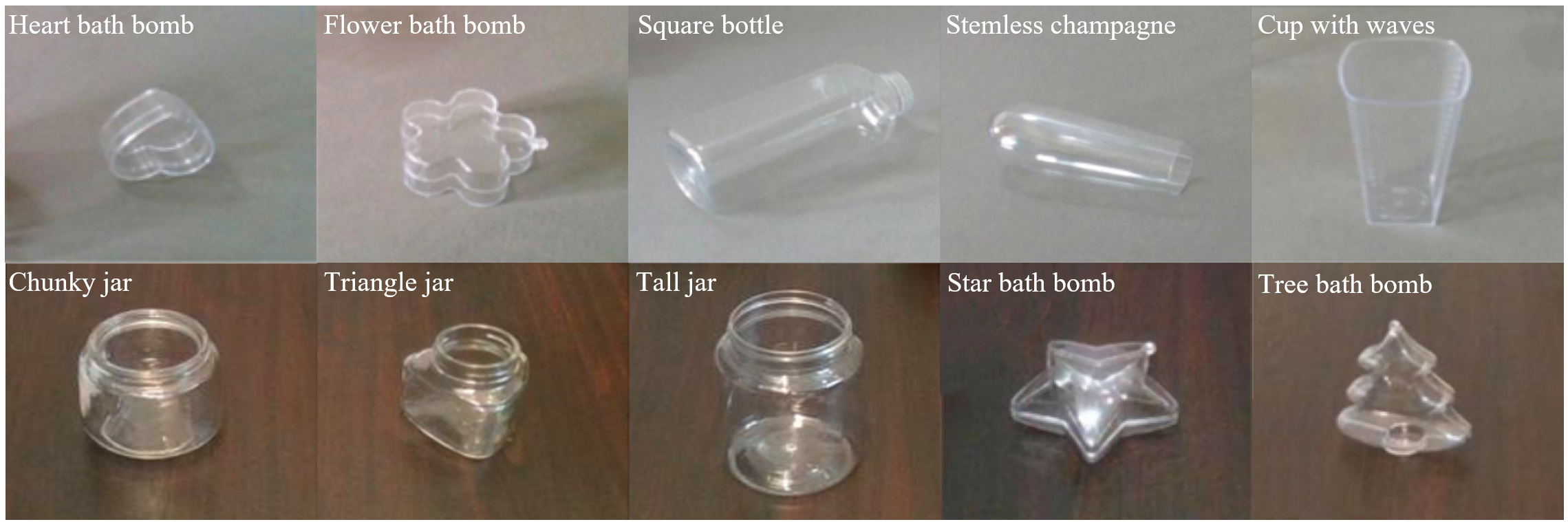}}
		\vspace{-1.0mm}
		\captionsetup{font={small}}
		\caption{Examples of transparent objects in ClearGrasp dataset}
		\vspace{-4.0mm}
		\label{Fig:2}
	\end{figure}

	We conducted our experiments on a large-scale transparent object dataset called ClearGrasp \cite{ref4}, which contains 9 types of synthetic objects and 10 types of real-world objects with a total number of over 50k images. The objects are shown in Fig. \ref{Fig:2}. In this work, we follow the data divisions in Sajjan et al. \cite{ref4} by using 5 known synthetic objects for training, and 5 overlapping real-world objects for testing. To verify the generalizability, 4 novel synthetic objects, and 5 novel real-world objects are used for testing.
	
	\captionsetup[table]{
		labelsep=newline,%换行
		singlelinecheck=true,%居左
	}
	\captionsetup[table]{labelformat=simple, labelsep=newline, textfont=sc,justification=centering}
	\captionsetup{font={scriptsize}}
	\begin{table}[h!]
		\vspace{2.4mm}
		\renewcommand{\arraystretch}{1.28}
		\setlength{\tabcolsep}{2.75mm}{
			\begin{center}
				\caption{THE PERFORMANCE OF THE APPROACHES ON DEPTH COMPLETION.}
				\begin{tabular}{c|cccccc} % <-- Alignments: 1st column left, 2nd middle and 3rd right, with vertical lines in between
					\hline
					\textbf{Method} & \textbf{RMSE}& \textbf{REL}& \textbf{MAE}& \textbf{$\delta_{1.05}$} & \textbf{$\delta_{1.10}$} & \textbf{$\delta_{1.25}$}
					\\
					\hline
					& \multicolumn{6}{c}{Cleargrasp Syn-known} \\
					\hline 
					CG & 0.041 & 0.055 & 0.031 & 69.43 & 89.17 & 96.74\\
					DG & 0.037 & 0.037 & 0.030 & 75.19 & 92.97 & 98.79\\
					DFNet & 0.018 & 0.023 & 0.013 & 88.95 & 97.57 & 99.92\\
					LIDF & 0.018 & 0.022 & 0.013 & 89.19 & 96.44 & 99.47\\
					FDCT & \textbf{0.015} & \textbf{0.020} & \textbf{0.012} & \textbf{90.53} & \textbf{98.21} & \textbf{99.99}\\
					Ours(S) & 0.019 & 0.033 & 0.017 & 86.43 & 95.18 & 99.01\\
					Ours(T)& 0.018 & 0.027 & 0.015 & 87.01 & 96.42 & 99.57 \\
					\hline
					& \multicolumn{6}{c}{Cleargrasp Syn-Novel} \\
					\hline 
					CG & 0.041 & 0.071 & 0.035 & 42.95 & 80.04 & 98.10\\
					DG & 0.039 & 0.062 & 0.032 & 51.86 & 82.14 & 98.32\\
					DFNet & 0.032 & 0.051 & 0.027 & 62.59 & 84.37 & 98.39\\
					LIDF & 0.038 & 0.057 & 0.031 & 58.64 & 82.03 & 98.59\\
					FDCT &0.025 & 0.040 & 0.021 & 71.66 & 92.95 & 99.64\\
					Ours(S) & 0.024 & 0.040 & 0.020 & 72.48 & 91.23 & 99.32\\
					Ours(T)& \textbf{0.021} & \textbf{0.035} & \textbf{0.018} & \textbf{77.56} & \textbf{93.83} & \textbf{99.68} \\
					\hline
					& \multicolumn{6}{c}{Cleargrasp Real-known} \\
					\hline 
					CG & 0.039 & 0.053 & 0.029 & 70.23 & 86.98 & 97.25\\
					DG & 0.031 & 0.039 & 0.021 & 74.69 & 89.73 & 97.35\\
					DFNet & 0.068 & 0.107 & 0.059 & 32.42 & 56.88 & 91.47\\
					LIDF & 0.039 & 0.050 & 0.030 & 69.51 & 85.03 & 96.44\\
					FDCT  & 0.065 & 0.103 & 0.057 & 33.08 & 59.81 & 91.70\\
					Ours(S) & 0.032 & 0.044 & 0.025 & 71.33 & 87.38 & 98.91\\
					Ours(T)& \textbf{0.021} & \textbf{0.032} & \textbf{0.018} & \textbf{80.78} & \textbf{94.91} & \textbf{99.56} \\
					\hline
					& \multicolumn{6}{c}{Cleargrasp Real-Novel} \\
					\hline 
					CG & 0.028 & 0.040 & 0.022 & 79.18 & 92.46 & 98.19\\
					DG & 0.022 & 0.033 & 0.017 & 82.37 & 93.46 & 98.48\\
					DFNet & 0.051 & 0.088 & 0.046 & 31.23 & 64.66 & 97.77\\
					LIDF & 0.035 & 0.056 & 0.030 & 54.37 & 82.70 & 98.27\\
					FDCT &0.043 & 0.073 & 0.038 & 39.42 & 75.54 & 99.09\\
					Ours(S) & 0.021 & 0.030 & 0.016 & 83.77 & 96.02 & 99.37\\
					Ours(T)& \textbf{0.020} & \textbf{0.028} & \textbf{0.015} & \textbf{84.12} & \textbf{96.06} & \textbf{99.43} \\
					\hline 
				\end{tabular}
				\label{Table:1}
		\end{center}}
		
		\vspace{-7.0mm}
	\end{table}
	\begin{figure*}[htbp]
		\vspace{1.0mm}
		\centering
		\subfloat[Visualization of the student network variants, where STU\_Alone represents without using teacher network,  STU\_CFCM\_DL represents without using CFCM and distillation loss, STU\_CFCM represents without using consistent feature correlation module (CFCM), STU\_DL represents without using distillation loss, and Student represents the complete student network.]{
			\includegraphics[width=0.94\linewidth]{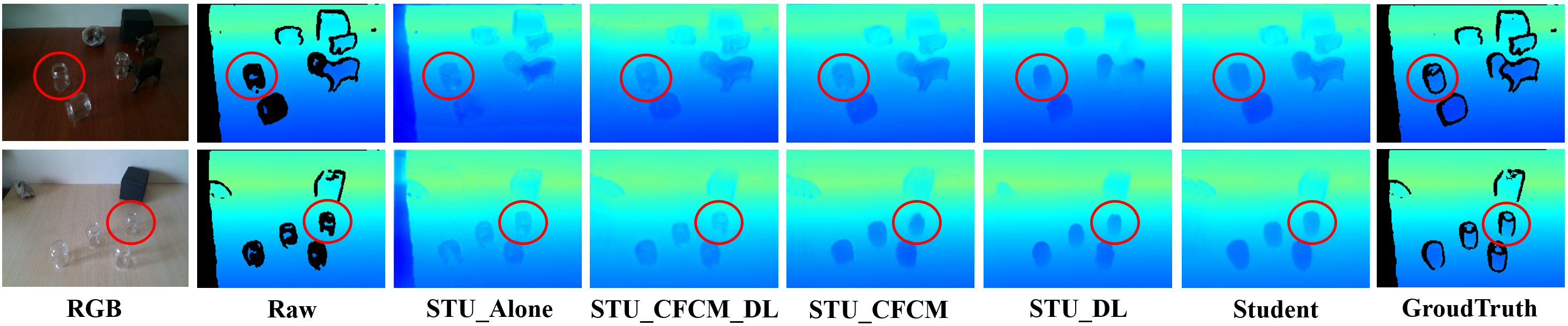}}
		\\
		\vspace{-3.0mm}
		\subfloat[Visualization of the teacher network variants, where TCH\_{}SA{}(Depth) and TCH\_{}SA{}(RGB) denote using self-attention mechanisms based on depth maps and RGB images respectively to replace the position correlation block (PCB). Teacher denotes the complete teacher network.]{
			\includegraphics[width=0.94\linewidth]{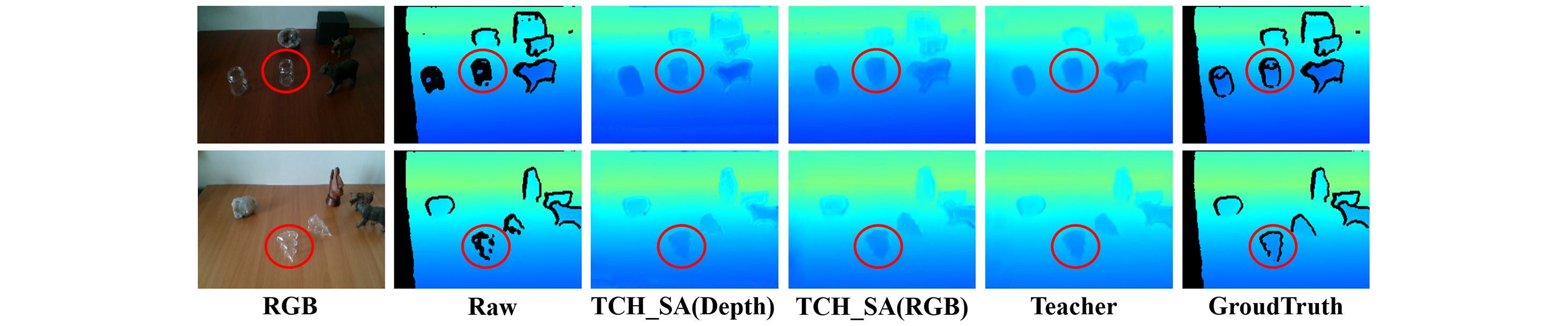}}\\
		\vspace{-0.9mm}
		\captionsetup{font={small}}
		\caption{Visualization of the teacher and student network variants, where the generated details are highlighted with red circles.}
		\vspace{-4.0mm}
		\label{Fig:4}
	\end{figure*}
	\subsection{Evaluation Metrics}
	Following previous works  \cite{ref4,ref5}, we adopt four common evaluation metrics including the root mean square error (RMSE), absolute relative difference (REL), mean absolute error (MAE), and accuracy with threshold $\delta_t$. Particularly, the threshold $ t $ is set to 1.05, 1.10, and 1.25, where 1.05 is the strictest requirement for evaluating depth completion. Note that all these metrics are calculated based on the transparent object areas.
	
	\subsection{Baselines}
	We compare our teacher network (denoted as \textbf{Our(T)}) and student network (denoted as \textbf{Our(S)}) with recent deep learning methods as follows:
	
	\textbf{ClearGrasp (CG) \cite{ref4}}, extracted multiple visual features from RGB-D images to infer the depths of transparent objects.
	
	\textbf{DepthGrasp (DG) \cite{ref5}}, designed a self-attentive adversarial network with spectral residual blocks to capture the geometric information and generate the missing points.
	
	\textbf{TransCG (DFNet) \cite{ref28}}, proposed an end-to-end depth completion network that uses U-Net architecture to extract fused features and outputs a refined depth map.
	
	\textbf{Local Implicit Function (LIDF) \cite{ref31}}, leveraged a local implicit neural representation based on ray-voxel pairs to interface the depth maps.
	
	\textbf{Fast Depth Completion (FDCT) \cite{ref29}}, introduced a fusion branch and cross-layer shortcuts to capture local information and designed a loss function to optimize the training process.
	\subsection{Performance of the Approaches on Depth Completion}
	Table \ref{Table:1} summarizes the performance of different approaches on depth completion, from which we have some observations. 1) CG and DG achieve stable results in both synthetic and real-world environments, manifesting that the additional geometrics features can provide extra structural information in reducing the gap between synthetic data and real-world data. However, the direct concatenation of different visual features limits the performance of the methods. 2) DFNet and FDCT achieve excellent results on the synthetic known objects, but they perform poorly on real-world objects, potentially due to the architecture of CNNs that only consider the local information and are more likely to get overfitting. 3) LIDF performs excellent results in both synthetic and real-world environments, however, the offset ray-voxel pairs will affect the surrounding depth values, leading to regional bias. 4) Our teacher network achieves a competitive result on synthetic known objects and outperforms other methods in synthetic novel objects, real-known objects and real-novel objects, demonstrating the effectiveness and generalizability of our method. In particular, our network has achieved a maximum of 6$\%$ improvement on real known objects, which is beneficial to the architecture of the transformer network for capturing geometric structure and position correlation block for correlating positional information. 5) Although there is a gap between the student network and the teacher network, the student network still achieves promising results on the novel objects. We believe it is because the teacher network effectively distills the structure and the edge information from the RGB images and depth maps and successfully teaches them to the student network.
	
	In Fig. \ref{Fig:3}, we visualize some examples of completed depth maps and highlight the generated details with red circles. From the figures we can observe that both our teacher network and student network can complete the depth map with clear shape and contour, but the student network still needs to improve the completion of object boundaries.
	
	\subsection{Speeds of the Approaches on Depth Completion}
	\vspace{0.2em}
	To simulate the real-world deployment, we use real novel objects for testing the time cost. Table \ref{Table:2} lists the time cost and parameters size of different approaches, from which we can observe that FDCT uses the least time to complete the depth map, however, its performance is not promising. Although our teacher network can achieve the best results in completing the depth map, it sacrifices twice as much time. Unlike the previous two networks, our student network achieves an ideal balance in time and accuracy, where it only sacrifices a little accuracy to acquire higher efficiency, reaching the speed of 48 FPS.

	\begin{table}[h!]
		\renewcommand{\arraystretch}{1.35}
		\vspace{2.5mm}
		\setlength{\tabcolsep}{1.18mm}{
			\begin{center}
				\caption{THE SPEED OF THE APPROACHES ON DEPTH COMPLETION.}
				\begin{tabular}{c|cccccccc} % <-- Alignments: 1st column left, 2nd middle and 3rd right, with vertical lines in between
					\hline
					\textbf{Method} & \textbf{RMSE}& \textbf{REL}& \textbf{MAE}& \textbf{$\delta_{1.05}$} & \textbf{$\delta_{1.10}$} & \textbf{$\delta_{1.25}$}
					& \textbf{Time} & \textbf{Size}\\
					\hline 
					CG & 0.028 & 0.040 & 0.022 & 79.18 & 92.46 & 98.19  & 1.9823s & 934MB\\
					DG & 0.022 & 0.033 & 0.017 & 82.37 & 93.46 & 98.48 & 2.1255s & 987MB\\
					LIDF & 0.035 & 0.054 & 0.029 & 55.06 & 82.75 & 98.79 & 0.0221s & 251MB\\
					DFNet & 0.051 & 0.088 & 0.046 & 31.23 & 64.66 & 97.77  & 0.0203s & 5.2MB\\
					FDCT &0.043 & 0.073 & 0.038 &39.42 & 75.54 & 99.09 & \textbf{0.0129s} & \textbf{4.8MB}\\
					Ours(S) & 0.021 & 0.030 & 0.016 & 83.77 & 96.02 & 99.37 & 0.0208s & 11.3MB\\
					Ours(T) & \textbf{0.020} & \textbf{0.028} & \textbf{0.015} & \textbf{84.12} & \textbf{96.06} & \textbf{99.43} & 0.0415s & 315MB\\
					\hline 
				\end{tabular}
				\label{Table:2}
		\end{center}}
		\vspace{-1.8em}
	\end{table}

	\subsection{Effect of Different Sets of Layers}
	As one of the important hyperparameters, the number of layers plays an important role in the results. Thus, in Table \ref{Table:5} we summarize the performance by choosing different sets of layers. Note that we use Teacher\_X and Student\_X to denote different types of networks, where X denotes the number of layers. From the table, we can notice that when the number of layers is set to 4, the network can achieve the best results. While using fewer or more layers, the performance will drop a little bit. We believe that it is because when using fewer sets of encoders, the network cannot further extract the details from the RGB images and depth maps, leading to lower accuracy. While using more encoders, the features will be compressed very small, where the spatial structure of the object will be violated. Moreover, the small size of features will increase the difficulties for the network to recover the depth maps.

	\begin{table}[h!]
		\renewcommand{\arraystretch}{1.35}
		\vspace{-1mm}
		\setlength{\tabcolsep}{2.75mm}{
			\begin{center}
				\vspace{1.8mm}
				\caption{EFFECT OF DIFFERENT SETS OF LAYERS.}
				\begin{tabular}{c|cccccc} % <-- Alignments: 1st column left, 2nd middle and 3rd right, with vertical lines in between
					\hline
					\textbf{Method} & \textbf{RMSE}& \textbf{REL}& \textbf{MAE}& \textbf{$\delta_{1.05}$} & \textbf{$\delta_{1.10}$} & \textbf{$\delta_{1.25}$}\\
					\hline 
					Teacher\_3 & 0.029 & 0.045 & 0.024 & 65.02 & 90.40 & 99.02 \\
					Teacher\_4 & \textbf{0.020} & \textbf{0.028} & \textbf{0.015} & \textbf{84.12} & \textbf{96.06} & \textbf{99.43} \\
					Teacher\_5 & 0.024 & 0.038 & 0.021 & 73.18 & 94.86 & 99.13 \\
					\hline 
					Student\_3 & 0.031 & 0.051 & 0.028 & 54.07 & 87.16 & 98.71 \\
					Student\_4 & \textbf{0.021} & \textbf{0.030} & \textbf{0.016} & \textbf{83.77} & \textbf{96.02} & \textbf{99.37}  \\
					Student\_5 & 0.027 & 0.046 & 0.024 & 63.00 & 92.73 & 99.02 \\
					\hline 
				\end{tabular}
				\label{Table:5}
		\end{center}}
		
		\vspace{-1.8em}
	\end{table}
	
	\subsection{Ablation Study}
	We design a series of ablation experiments to verify each component. For the teacher network, we adopt self-attention mechanisms based on RGB images and depth maps respectively to replace the position correlation block (PCB) and denote them as TCH\_{}SA{}(RGB) and TCH\_{}SA{}(Depth), respectively. For the student network, we denote the student network without consistent feature correlation module (CFCM) as STU\_CFCM, the student network without distillation loss as STU\_DL, the student network without CFCM and distillation loss as STU\_CFCM\_DL, and student network without teacher network as STU\_Alone. Note that we use common loss RMSE to replace the distillation loss and adopt UNet as our baseline model. Additionally, we further verify the effect of each distillation loss.
	
	Table \ref{Table:3} shows the experimental results, from which we can find that our PCB in the teacher network achieves a maximum 18.2$\%$ improvement compared with the single input of self-attention, illustrating that the correlation of features can facilitate depth completion. In the student network, the employment of CFCM and distillation loss obtain a certain improvement in accuracy, where the CFCM has a greater impact on the depth completion because it fuses the two different types of features and decides the entire quality of the depth completion, while distillation loss only plays a part in optimizing the generation of edge and object structure. 
	
	Intuitively, we visualize some examples of the completed depth maps generated by the ablation experiments and highlight the generated details with red circles. The results are shown in Fig. \ref{Fig:4}. From Fig. \ref{Fig:4}(a) we can observe that our position correlation block can better deal with the complex shapes objects owing to the correspondence established between RGB images and depth maps. From Fig. \ref{Fig:4}(b) we can notice that the adoption of CFCM has a better result on the overall depth completion and the usage of distillation loss can improve the generation of the object structure and contour.
	\begin{table}[h!]
		\renewcommand{\arraystretch}{1.35}
		\setlength{\tabcolsep}{1.835mm}{
			\begin{center}
				\vspace{1.8mm}
				\caption{THE ABLATION EXPERIMENT OF STUDENT AND TEACHER NETWORK.}
				\begin{tabular}{c|cccccc} % <-- Alignments: 1st column left, 2nd middle and 3rd right, with vertical lines in between
					\hline
					\textbf{Method} & \textbf{RMSE}& \textbf{REL}& \textbf{MAE}& \textbf{$\delta_{1.05}$} & \textbf{$\delta_{1.10}$} & \textbf{$\delta_{1.25}$}
					\\
					\hline
					& \multicolumn{6}{c}{Teacher} \\
					\hline 
					TCH\_{}SA\_{}(RGB) & 0.023 & 0.037 & 0.019 & 75.70 & 92.83 & 99.41\\
					TCH\_{}SA\_{}(Depth) & 0.028 & 0.046 & 0.023 & 65.92 & 87.59 & 99.20\\
					Teacher & \textbf{0.020} & \textbf{0.028} & \textbf{0.015} & \textbf{84.12} & \textbf{96.06} & \textbf{99.43}\\
					\hline
					& \multicolumn{6}{c}{Student} \\
					\hline 
					STU\_{}Alone & 0.032 & 0.056 & 0.032 & 59.91 & 92.30 & 98.53\\
					STU\_{}CFCM\_{}DL & 0.026 & 0.036 & 0.025 & 69.65 & 93.65 &  98.89\\
					STU\_{}CFCM & 0.024 & 0.037 & 0.021 & 74.06 & 93.97 & 99.06 \\
					STU\_{}DL & 0.023 & 0.034 & 0.019 & 78.23 & 94.76 & 99.20\\
					Student & \textbf{0.021} & \textbf{0.030} & \textbf{0.016} & \textbf{83.77} & \textbf{96.02} & \textbf{99.37}\\
					\hline
				\end{tabular}
				\label{Table:3}
		\end{center} }
		
		\vspace{-1.7em}
	\end{table}
	
	Table \ref{Table:5} illustrates the effect of each distillation loss, from which we can notice that the distance loss plays a critical role in the performance as it computes the point-wise errors and decides the overall completion. The structure loss and edge loss determine the object structure and boundary details respectively, which improve the results to a certain extent. The combination of loss functions achieves the best results since it guarantees depth completion in overall effect and details.

	\begin{table}[h!]
	\renewcommand{\arraystretch}{1.35}
	\setlength{\tabcolsep}{1.06mm}{
		\begin{center}
			\caption{THE ABLATION EXPERIMENT OF DISTILLATION LOSS.}
			\begin{tabular}{c|c|c|cccccc} % <-- Alignments: 1st column left, 2nd middle and 3rd right, with vertical lines in between
				\hline
				\textbf{Distance} & \textbf{Structural} &\textbf{Edge} & \textbf{RMSE}& \textbf{REL}& \textbf{MAE}& \textbf{$\delta_{1.05}$} & \textbf{$\delta_{1.10}$} & \textbf{$\delta_{1.25}$} \\
				\hline 
				\checkmark & ~ & ~ & 0.023 & 0.033 & 0.018 & 79.23 & 95.21 & 99.27 \\
				~ & \checkmark & ~ & 0.024 & 0.034 & 0.019 & 78.61 & 95.28 & 99.29 \\
				~ & ~ & \checkmark & 0.026 & 0.038 & 0.021 & 75.89 & 93.58 & 99.23 \\ 
				\checkmark & \checkmark & ~ & 0.021 & 0.031 & 0.016 & 82.31 & 95.42 & 99.37 \\ 
				\checkmark & ~ & \checkmark & 0.022 & 0.033 & 0.017 & 80.23 & 95.23 & 99.36 \\ 
				~ & \checkmark & \checkmark & 0.022 & 0.032 & 0.017 & 79.50 & 94.66 & 99.33 \\ 
				\checkmark & \checkmark & \checkmark & \textbf{0.021} & \textbf{0.030} & \textbf{0.016} & \textbf{83.77} & \textbf{96.02} & \textbf{99.37} \\
				\hline 
			\end{tabular}
			\label{Table:5}
		\end{center}}
	\vspace{-1.8em}
	\end{table}

	\subsection{Robot Grasping}
	\begin{figure}[htbp]
		\centering
		\subfloat[Real Novel objects]{
			\includegraphics[width=0.98\linewidth]{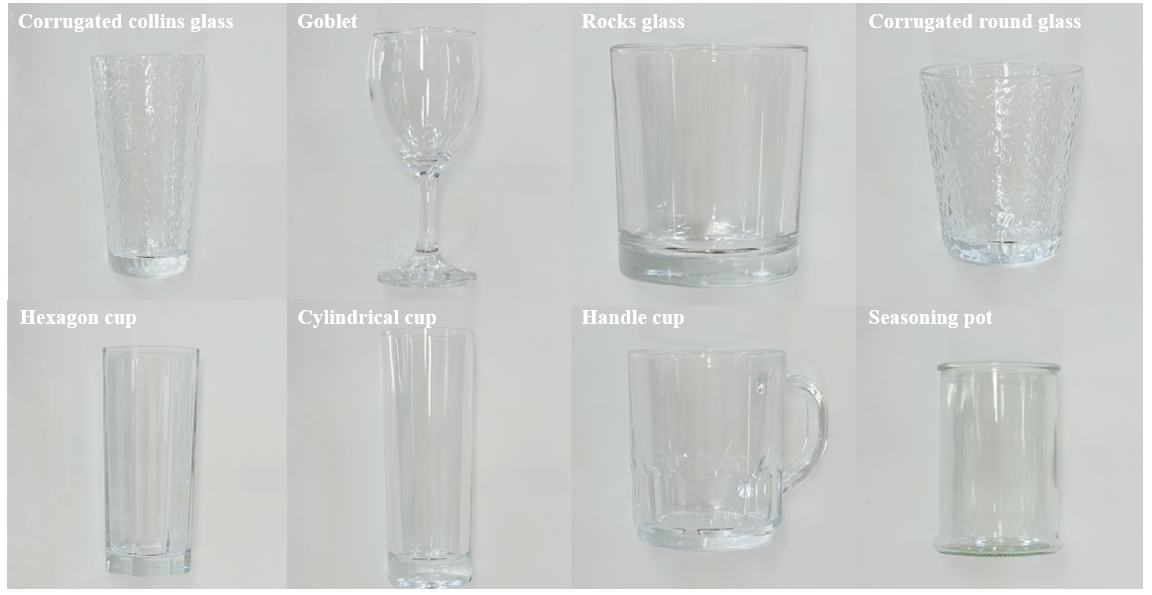}}
		\\
		\vspace{-2.5mm}
		\subfloat[Real robot Grasping]{
			\includegraphics[width=0.98\linewidth]{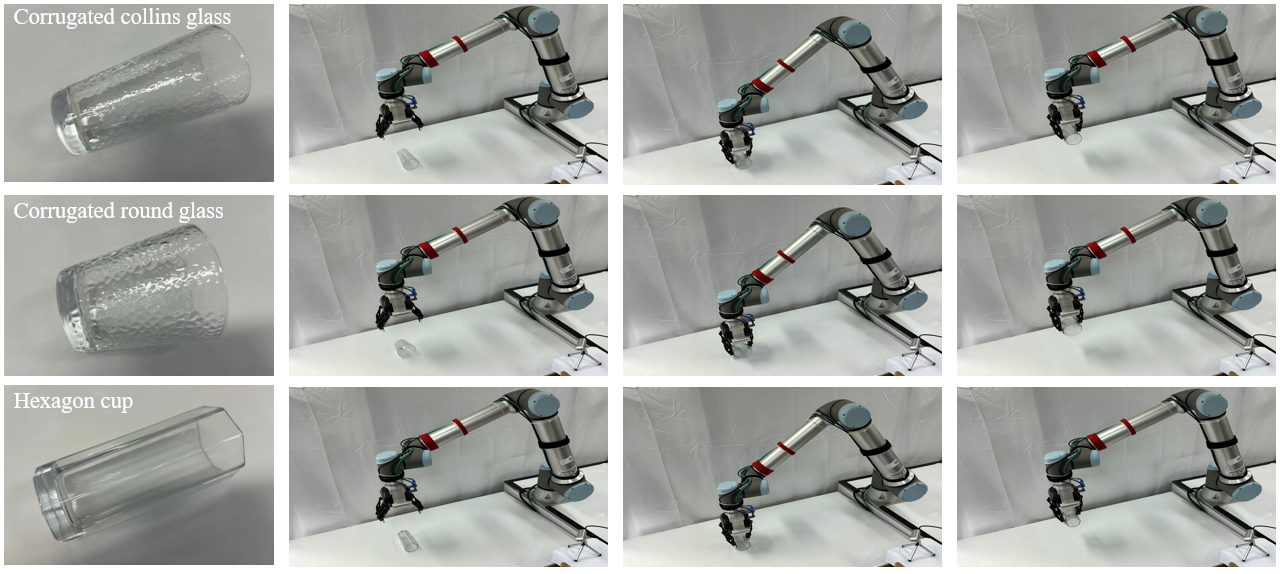}}
		\
		\captionsetup{font={small}}
		\vspace{-1.5mm}
		\caption{UR10e grasping the transparent objects.}
		\vspace{-1.5mm}
		\label{Fig:5}
	\end{figure}

	In order to demonstrate the applicability of our network in practical tasks, we deployed our system on a single-arm robot UR10e to grasp real-world transparent objects. Specifically, we choose the GR-CNN \cite{ref30}, which was verified by the previous work \cite{ref5}, as our grasping method. Fig. \ref{Fig:5}(a) presents 8 novel transparent objects that do not appear in the ClearGrasp dataset for grasping, including  “Corrugated collins glass”, “Goblet”, “Rocks glass”, “Corrugated round glass”, “Hexagon cup”, “Handler cup”, “Cylindrical cup” and “Seasoning pot”. For each object, the robot will attempt to grasp it 10 times, and only when the object is held for more than 10 seconds will be considered successful. 
	
	Table \ref{Table:4} lists the success rate of grasping, from which we can observe that our proposed network can tremendously improve the success rate compared with using raw depth maps and outperform the state-of-the-art method DepthGrasp in real-world grasping. Fig. \ref{Fig:5}(b) shows a few snapshots of UR10e grasping transparent objects, from which we can see that the robot can successfully hold the transparent objects, demonstrating the effectiveness of our proposed method.
	
	\begin{table}[h!]
		\renewcommand{\arraystretch}{1.35}
		\setlength{\tabcolsep}{3.23mm}{ %12可随机设置，调整到适合自己的大小为止
			\vspace{1.8mm}
			\caption{REAL OBJECTS GRASPING.}
			\begin{tabular*}{\linewidth}{c|cccc} % <-- Alignments: 1st column left, 2nd middle and 3rd right, with vertical lines in between
				\hline
				\textbf{Object} & \textbf{Raw}& \textbf{DG}
				& \textbf{Ours(S)} & \textbf{Ours(T)}\\
				\hline
				Corrugated collins glass & 3/10 & 8/10 & 8/10 & 8/10 \\
				Goblet & 1/10 & 8/10 & 8/10 & 9/10 \\
				Rocks glass & 3/10 & 7/10 & 8/10 & 8/10\\
				Corrugated round glass & 2/10 & 8/10 & 8/10 & 8/10 \\
				Hexagon cup & 5/10 & 9/10 & 7/10 & 9/10 \\
				Cylindrical cup & 5/10 & 8/10 & 9/10 &  9/10 \\
				Handler cup & 2/10 & 7/10 & 7/10 &  8/10\\
				Seasoning pot & 4/10 & 7/10 & 8/10 & 8/10\\
				\hline
				Success rate($\%$) & \textbf{36.25} & \textbf{77.50} & \textbf{78.75} & \textbf{83.75} \\
				\hline
			\end{tabular*}
			\label{Table:4}
		}
		\vspace{-1.7mm}
	\end{table}
	
	\section{Conclusion}
	In this paper, we propose a novel distillation network called DistillGrasp for completing the depth of transparent objects. Specifically, in the teacher network, we proposed a position correlation block to search for the positional correspondence between RGB images and depth maps. For the student network, we designed an efficient consistent feature correlation module to capture the pairwise relationship based on reliable positional information. In order to transfer the overall knowledge from the teacher branch to the student branch, we present a distillation loss that takes distance loss, object structure, and edge information into consideration. Extensive experiments demonstrated that our teacher network achieves state-of-the-art performance on accuracy and generalization while the student network achieves ideally balanced results in terms of efficiency and speed. Moreover, the deployment of our system on a robot effectively verifies the applicability of our system.

	\section*{Acknowledgment}
	This work is supported by the National Natural Science Foundation of China (No. 91748107), the Special Research Fund (BOF) of Hasselt University (No. BOF23DOCBL11), the foundation of State Key Laboratory of Public Big Data(No. PBD2023-11), the Guangdong Innovative Research Team Program (No. 2014ZT05G157). Chen Junhong was sponsored by the China Scholarship Council (No. 202208440309).

	\ifCLASSOPTIONcaptionsoff
	\newpage
	\fi
	\small
	\bibliographystyle{ieeetr}

\begin{thebibliography}{40}
		\bibitem{ref1}
		J. Weibel, P. Sebeto, S. Thalhammer, and M. Vincze, ``Challenges of Depth Estimation for Transparent Objects,'' International Symposium on Visual Computing, pp. 277-288, 2023.
		\bibitem{ref4}
		S. Sajjan, M. Matthew, M. Pan, N. Ganesh, J. Lee, A. Zeng, and S. Song, ``ClearGrasp:3D shape estimation of transparent objects for manipulation,'' IEEE International Conference on Robotics and Automation (ICRA), pp. 1-13, 2020.
		\bibitem{ref5}
		Y. Tang, J. Chen, Z. Yang, Z. Lin, Q. Li, and W. Liu, ``Depthgrasp: Depth completion of transparent objects using selfattentive adversarial network with spectral residual for grasping,'' IEEE International Conference on Intelligent Robots and Systems (IROS), pp. 5710-5716,2021.
		\bibitem{ref6}
		K. Chen, S. Wang, B. Xia, D. Li, Z. Kan, and B. Li, ``TODETrans: Transparent Object Depth Estimation with Transformer,'' IEEE International Conference on Robotics and Automation (ICRA), pp. 4880-4886, 2023.
		\bibitem{ref7}
		Y. Hong, J. Chen, Y. Cheng, Y. Han, F. Van Reeth, L. Claesen, and W. Liu, ``Cluedepth grasp: Leveraging positional clues of depth for completing depth of transparent objects,'' Frontiers in Neurorobotics, vol. 16, 2022.
		\bibitem{ref8}
		J. Jiang, G. Cao, J. Deng, T. -T. Do and S. Luo, "Robotic Perception of Transparent Objects: A Review," in IEEE Transactions on Artificial Intelligence, vol. 5, no. 6, pp. 2547-2567, 2023.
		\bibitem{ref9}
		U. Klank, D. Carton, and M. Beetz, ``Transparent object detection and reconstruction on a mobile platform,'' IEEE International Conference on Robotics and Automation (ICRA), 2011, pp. 5971–5978.
		\bibitem{ref11}
		J. Ichnowski, Y. Avigal, J. Kerr, and K. Goldberg, ``Dex-nerf: Using a neural radiance field to grasp transparent objects,'' Conference on Robot Learning (CoRL), 2021.
		\bibitem{ref12}
		J. Kerr, L. Fu, H. Huang, J. Ichnowski, M. Tancik, Y. Avigal, A. Kanazawa, and K. Goldberg, ``Evo-nerf: Evolving nerf for sequential robot grasping,'' Conference on Robot Learning (CoRL), 2022.
		\bibitem{ref13}
		Y. Qian, M. Gong, and Y. H. Yang, ``3d reconstruction of transparent objects with position-normal consistency,'' in Proc. IEEE/CVFInternational Conference on Pattern Recognition(ICPR), pp. 4369–4377, 2016.
		\bibitem{ref19}
		Hinton, Geoffrey, Oriol Vinyals, and Jeff Dean, ``Distilling the knowledge in a neural network,'' arXiv preprint arXiv:1503.02531, 2015.
		\bibitem{ref32}
		Z. Chen, Z. Li, S. Zhang, L. Fang, Q. Jiang, and F. Zhao, "Bevdistill: Cross-modal bev distillation for multi-view 3d object detection," International Conference on Learning Representations (ICLR), 2023.
		\bibitem{ref33}
		Y. Hong, H. Dai, and Y. Ding, "Cross-modality knowledge distillation network for monocular 3d object detection," in European Conference on Computer Vision. Springer, 2022, pp. 87–104.
		\bibitem{ref34}
		J. Cen, S. Zhang, Y. Pei, K. Li, H. Zheng, M. Luo, Y. Zhang, and Q. Chen, "Cmdfusion: Bidirectional fusion network with cross-modality knowledge distillation for lidar semantic segmentation," IEEE Robotics and Automation Letters, vol. 9, no. 1, pp. 771–778, 2023.
		\bibitem{ref35}
		F. Jiang, H. Gao, S. Qiu, H. Zhang, R. Wan, and J. Pu., "Knowledge Distillation from 3D to Bird’s-Eye-View for LiDAR Semantic Segmentation," in 2023 IEEE International Conference on Multimedia and Expo (ICME), Brisbane, Australia, 2023 pp. 402-407.
		\bibitem{ref24}
		W. Yuan, X. Gu, Z. Dai, S. Zhu, and P. Tan, ``Neural Window Fully-connected CRFs for Monocular Depth Estimation,''  IEEE Conference on Computer Vision and Pattern Recognition (CVPR), pp. 3906-3915, 2022.
		\bibitem{ref26}
		Z. Liu, Y. Lin, Y. Cao, H. Hu, Y. Wei, Z. Zhang, S. Lin, and B. Guo, ``Swin transformer: Hierarchical vision transformer using shifted windows,'' IEEE/CVF International Conference on Computer Vision (ICCV),  pp. 10012-10022, 2021.
		\bibitem{ref27}
		D. Eigen, C. Puhrsch, and R. Fergus, ``Depth map prediction from a single image using a multi-scale deep network,'' Neural Information Processing Systems(NIPS), pp. 2366-2374, 2014.
		\bibitem{ref28}
		H. Fang, H. -S. Fang, S. Xu and C. Lu, ``TransCG: A Large-Scale Real-World Dataset for Transparent Object Depth Completion and a Grasping Baseline,'' IEEE Robotics and Automation Letters, vol. 7, no. 3, pp. 7383-7390, 2022.	
		\bibitem{ref31}
		L. Zhu, A. Mousavian, Y. Xiang, H. Mazhar, J. Eenbergen, S. Debnath, and D. Fox, ``RGB-D Local Implicit Function for Depth Completion of Transparent Objects,'' IEEE/CVF Conference on Computer Vision and Pattern Recognition (CVPR),  pp. 4647-4656, 2021.
		\bibitem{ref29}
		T. Li, Z. Chen, H. Liu, and C. Wang, ``FDCT: Fast Depth Completion for Transparent Objects,''  IEEE Robotics and Automation Letters, vol. 8, no. 9, pp. 5823-5830, 2023.
		\bibitem{ref30}
		S. Kumra, S. Joshi, and F. Sahin, ``Antipodal robotic grasping using generative residual convolutional neural network,'' IEEE International Conference on Intelligent Robots and Systems (IROS), pp. 9626-9633, 2020.
	\end{thebibliography}

\end{document}